\documentclass[a4paper]{article}

\usepackage{amsmath}
\usepackage{amsthm}
\usepackage{amssymb}
\usepackage{bm}
\usepackage{hyperref}
\usepackage{dsfont}
\usepackage{bbm}
\usepackage{graphicx}

\usepackage{setspace}

\usepackage{multirow,booktabs,setspace,caption}
\usepackage{xcolor, colortbl}
\definecolor{lightgray}{gray}{0.70}
\newcommand{\zero}{\textcolor{lightgray}{0.00}}

\usepackage{geometry}
\usepackage{algorithm}
\usepackage{algpseudocode}

\graphicspath{{./figures/}}
\usepackage[backend=biber,style=vancouver,sorting=none,maxnames=3,minnames=1]{biblatex}
\DeclareFieldFormat{labelnumberwidth}{#1.}
\DeclareNameAlias{author}{last-first}
\DeclareNameAlias{editor}{last-first}

\DefineBibliographyStrings{english}{andothers = {et al}}

\providecommand{\keywords}[1]
{
  \small	
  \textit{Keywords---} #1
}

\begin{document}

\title{Ensemble Machine Learning and Statistical Procedures for Dynamic Predictions of Time-to-Event Outcomes}
\author{Nina van Gerwen$^{1,2}$, Sten Willemsen$^{1,2}$, Bettina E. Hansen$^{1,2}$, Christophe Corpechot$^{3}$, \\ Marco Carbone$^{4}$, Cynthia Levy$^{5}$, Maria-Carlota Londo\~no$^{6}$, Atsushi Tanaka$^{7}$, \\Palak Trivedi$^{8}$, Alejandra Villamil$^{9}$, Gideon Hirschfield$^{10}$, Dimitris Rizopoulos$^{1,2}$}

\date{\small{
	$^1$Department of Biostatistics, Erasmus University Medical Center, the Netherlands\\
	$^2$Department of Epidemiology, Erasmus University Medical Center, the Netherlands\\
	$^3${Reference Center for Inflammatory Biliary Diseases and Autoimmune Hepatitis, European Reference Network on Hepatological Diseases (ERN Rare-Liver), Saint-Antoine Hospital, Assistance Publique - H\^opitaux de Paris; Inserm UMR\_S938, Saint-Antoine Research Center, Sorbonne University, Paris, France}\\
	$^4$Department of Medicine and Surgery, University of Milan Bicocca, Italy\\
	$^5$Miller School of Medicine, University of Miami, United States of America\\
	$^6$Viral, Toxic and Metabolic Hepatopathies, Hospital Cl\'inic de Barcelona, Spain\\
	$^7$Department of Medicine, Teikyo University School of Medicine, Japan\\
	$^8$Biomedical Research Unit \& Centre for Liver Research, University of Birmingham, United Kingdom\\
	$^{9}$Division of Gastroenterology \& Hepatology, Hospital Italiano de Buenos Aires, Argentina\\
	$^{10}$Centre for Liver Disease, Toronto General Hospital, Canada\\}	
	\today}

\maketitle
\small{\section*{Correspondence}}
Erasmus MC attn. Nina van Gerwen, Wytemaweg 12, 3015 CN, Rotterdam, the Netherlands
\small{\section*{Acknowledgements}}
\small{We would like to thank the Global PBC Study Group for granting us availability of their data for the current article.}
\section*{Statements and declarations}
\subsection*{Ethical considerations}
Ethical approval for the analysis and publication of the retrospectively obtained and anonymised data of the Global PBC Study Group was not required.
\subsection*{Consent to participate}
Not applicable.
\subsection*{Consent for publication}
Not applicable.
\subsection*{Declaration of conflicting interest}
The authors declared no potential conflicts of interest with respect to the research, authorship, and/or publication of this article.
\subsection*{Funding statement}
This work is funded by Stress in Action. The research project ‘Stress in Action’: www.stress-in-action.nl is financially supported by the Dutch Research Council and the Dutch Ministry of Education, Culture and Science (NWO gravitation grant number 024.005.010).
\subsection*{Data Availability Statement}
The data that support the findings of this study are available from the Global PBC Study Group. Restrictions apply to the availability of these data, which were used under license for this study. Data are available at https://www.globalpbc.com with the permission of the Global PBC Study Group.
%
%
\newpage
\begin{abstract} 
Dynamic predictions for longitudinal and time-to-event outcomes have become a versatile tool in precision medicine. Our work is motivated by the application of dynamic predictions in the decision-making process for primary biliary cholangitis patients. For these patients, serial biomarker measurements (e.g., bilirubin and alkaline phosphatase levels) are routinely collected to inform treating physicians of the risk of liver failure and guide clinical decision-making. Two popular statistical approaches to derive dynamic predictions are joint modelling and landmarking. However, recently, machine learning techniques have also been proposed. Each approach has its merits, and no single method exists to outperform all others. Consequently, obtaining the best possible survival estimates is challenging. Therefore, we extend the Super Learner framework to combine dynamic predictions from different models and procedures. Super Learner is an ensemble learning technique that allows users to combine different prediction algorithms to improve predictive accuracy and flexibility. It uses cross-validation and different objective functions of performance (e.g., squared loss) that suit specific applications to build the optimally weighted combination of predictions from a library of candidate algorithms. In our work, we pay special attention to appropriate objective functions for Super Learner to obtain the most optimal weighted combination of dynamic predictions. In our primary biliary cholangitis application, Super Learner presented unique benefits due to its ability to flexibly combine outputs from a diverse set of models with varying assumptions for equal or better predictive performance than any model fit separately.\\
\end{abstract}
\keywords{dynamic predictions, ensemble learning, joint models, landmarking, predictive accuracy}
\section{Introduction}
Dynamic predictions for longitudinal and time-to-event outcomes, where predictions are updated as more information is collected over time, have become a versatile tool in precision medicine for numerous diseases. \supercite{disease1, disease2, disease3, disease4} Our work is motivated by primary biliary cholangitis (PBC), a slow progressive and chronic autoimmune liver disease. There is currently no cure available for the disease, and a large proportion of patients have an inadequate response to the current clinical guidance treatment ursodeoxycholic (UDCA) therapy.\supercite{pbcnon} Thus, after diagnosis, PBC patients are routinely followed up for the remainder of their lives to track disease progression and identify patients at risk of liver failure (either an acute deterioration in liver function or the need for liver transplantation). For these patients, several liver-related biomarkers are routinely measured. Physicians then use patients' biomarker values to guide clinical decision-making, such as who should receive a liver transplantation or be more closely followed up. To help better make these critical decisions for PBC patients, a survival model that gives highly accurate dynamic time-to-event prediction is paramount. \\
\indent Currently, the two most commonly used procedures for dynamic time-to-event prediction are joint modelling\supercite{jm} and landmarking,\supercite{lm} though more methods exist (e.g., DynamicDeep-Hit,\supercite{DDH} a neural network based approach). However, ensuring that a single model or procedure provides the most accurate dynamic predictions can be challenging for several reasons. Foremost, researchers need to have a considerable amount of a-priori information on different aspects of model specification with joint modelling or landmarking, such as variable inclusion, model terms and functional forms to link the two outcomes. For example, to accurately predict disease progression in PBC patients, a researcher should know: (a) what risk factors, biomarkers and interactions to include in the model, (b) the trajectory over time of the different biomarkers (e.g., whether the biomarkers show linear trends or nonlinear trends over time within patients), and (c) how the various biomarkers are related to disease progression (e.g., whether current biomarker concentration in the patients' blood or the rate at which the biomarker concentration increases predicts disease progression). \\
\indent Commonly, researchers use standard model selection techniques (e.g., information criteria or significance tests) to answer these questions by choosing the model that displays the best fit. However, in the context we consider, joint models and landmarking cannot be directly compared because they are estimated using different likelihood functions. Furthermore, the likelihood of joint models is often dominated by the longitudinal outcome, and information criteria may not focus on optimising the predictive accuracy of the time-to-event outcome. Researchers can also use metrics of predictive accuracy (e.g., Brier Score (BS) or area-under-the-receiver operating curve (AUC)) to determine the best model. However, these metrics have been shown to not be sensitive enough to determine differences between models.\supercite{BS_bad, BS_bad2} Finally, it is worth noting that because we work with multiple outcomes the set of possible statistical models increases exponentially, further complicating model specification. \\
\indent We propose to use ensemble learning techniques in the setting of dynamic predictions to overcome the issues described above. In ensemble learning, the predictive performance of multiple models is combined to provide better and more flexible predictions than any single model. We extend the ensemble learning technique Super Learner\supercite{SL} (SL) in the context of dynamic prediction. Previously, Rizopoulos et al.\supercite{jm_sl} have shown how SL can combine multiple joint models to accommodate uncertainty in the association between the longitudinal and time-to-event outcome. However, they focused on combining exclusively joint models to accommodate different longitudinal trajectories and functional forms of the association between the two outcomes, and they paid less attention to different frameworks. Furthermore, joint models are known to be a computationally intensive method, and using solely joint models in an ensemble for SL explosively increases computational complexity. Lastly, joint models need to be fully pre-specified in covariate selection and model terms. Tanner et al.\supercite{keogh_sl} took another approach to SL for dynamic predictions. By discretising time, they showed how SL can combine binary classification algorithms to estimate dynamic predictions. Even though this results in a very flexible algorithm with data-driven covariate selection, discretising time brings the disadvantages of information loss and subjectivity in the size of the discrete time intervals. Both papers also focused on optimising predictive accuracy by minimising BS or expected predictive cross-entropy. Although SL has been extended to optimise the time-invariant AUC for binary classification problems,\supercite{auc} SL for dynamic predictions has not been extended to optimising the time-varying AUC.\supercite{tv_auc} (\textit{tv}-AUC) Therefore, our work builds upon these methods by combining the predictive performance of different dynamic prediction procedures in continuous time for the optimisation of previously used and new metrics. We also further investigate the performance of SL in varying settings of longitudinal and time-to-event outcomes. \\
\indent The remainder of the paper is organised as follows: Section 2 formally details the methods for SL with dynamic predictions. In Section 3, we employ SL for dynamic predictions in the Global PBC dataset. Section 4 introduces a simulation study that examines the performance of SL for dynamic predictions with varying forms of censoring. Lastly, Section 5 presents the findings and practical implications.
\section{Super Learner for dynamic predictions}
\indent Suppose we have a sample $\mathcal{D}_n = \{ T_i, \delta_i, \boldsymbol{\mathit{y}}_{i} ; i = 1, \ldots, n\}$ of size $n$ from our target population. Let $T_i^*$ denote the true event time for subject $i$. We denote $T_i = \text{min}(T_i^*, C_i)$ as the observed event time, where $C_i$ denotes the censoring time. The event indicator $\delta_i = \mathbbm{1} (T_i \le C_i)$, in which $\mathbbm{1}(\cdot)$ is an indicator function that equals 1 when the condition within holds and 0 otherwise, tells us whether individual $i$ was censored or experienced the event of interest. Finally, $\boldsymbol{\mathit{y}}_{i} = (\boldsymbol{\mathit{y}}_{i1}, \dots, \boldsymbol{\mathit{y}}_{iM})$ is a vector of $m = 1, \dots, M$ longitudinal outcomes of response- and subject-specific size $k_{m,i}$, where $\mathit{y}_{m,il}$ is the value of the $m$th longitudinal response at time points $t_{m,il}, l = 1, \ldots, k_{m,i}$. These time points need not be regular intervals and can differ per subject and longitudinal outcome. We assume right-censoring occurred where subjects are lost to follow-up. \\
\indent The objective is to make accurate survival predictions for a new subject \textit{j} from the target population. This subject should then also have sets of longitudinal measurements up to $t: \boldsymbol{\mathit{y}}_{j}(t)$, which concurrently implies survival of subject \textit{j} up to $t$. Therefore, we focus on survival predictions conditional on survival at $t$. Mathematically, we are interested in the survival probability at time $u > t$:
\begin{equation}\label{eq:dynpred}
 \pi_j(u \mid t) = \text{Pr}(T^*_j \ge u \mid T^*_j > t, \boldsymbol{\mathit{y}}_{j}(t), \mathcal{D}_n).
\end{equation}
This probability can then be updated whenever a new measurement for subject $j$ becomes available at $t^+ > t$ to obtain $\pi_j(u \mid t^+)$. Different methods to estimate \eqref{eq:dynpred} are discussed in Section 2.2. \\
\indent In SL, we assume that we have set $\mathcal{K} = \{\mathcal{M}_1, \dots, \mathcal{M}_K\}$ of $K$ base learners for $\mathcal D_n$ that can estimate \eqref{eq:dynpred}. Importantly, these base learners come from different modelling frameworks (e.g., landmarking, joint modelling, etc.) and vary in specification within a framework (e.g., different baseline covariates, different model terms, etc.). 
The process of SL works as follows. First, we split up $\mathcal{D}_n$ into $v = {1, \dots V}$ folds and obtain cross-validated estimates of $\pi_{i}(u \mid t)$ by fitting each base learner in $\mathcal{K}$ on $V - 1$ folds, and estimating predictions on the $v$-th fold excluded in model fitting. This is repeated for each fold. We end up with $K$ predictions for all $i \in \mathcal{R}(t)$, where $\mathcal{R}(t)$ denotes the set of individuals still at risk at time $t$ (i.e., $T_i > t$):
\begin{equation*}
\hat{\pi}_{i}^{v}(u \mid t, \mathcal{M}_k) = \text{Pr}(T_i^* \ge u \mid T_i^* > t, \mathcal{M}_k, \mathcal{D}_n^{-v}),
\end{equation*}
where $\mathcal{M}_k$ is base learner $k \in K$ and $\mathcal{D}_n^{-v}$ is $\mathcal{D}_n$ with individuals from the $v$-th fold removed. Then we assign weights $\omega_k(t)$ to the estimated cross-validated predictions. SL optimises predictive accuracy by estimating $\omega_k(t)$ such that an a-priori chosen loss function $\lambda(\cdot)$ is minimised. The result is an ensemble SL (eSL):
\begin{equation}\label{eq:esl}
\tilde{\pi}_i^v(u \mid t) = \sum_{k = 1}^K \hat{\omega}_k(t) \hat{\pi}_i^v(u \mid t, \mathcal{M}_k), \hspace{0.5cm} \text{for} \hspace{0.1cm} v = 1, \dots V,
\end{equation}
in which
\begin{equation*}
\hat{\omega}_k(t) = \underset{\omega_k(t)}{\text{argmin}} \biggr\{ \sum_{i \in \mathcal{R}(t)} \lambda \Bigr( \sum_{k = 1}^K \omega_k(t) \hat{\pi}_i^v(u \mid t, \mathcal{M}_k),\, T_i \Bigr) \biggr\}, 
\end{equation*}
subject to
\begin{equation*}
\omega_k(t) \ge 0 \ \wedge \ \sum_{k = 1}^{K} \omega_k(t) = 1.
\end{equation*}
Choosing an appropriate $\lambda(\cdot)$ that fits the research objective is important, as this will define the estimation of $\omega_k(t)$. We focus on three possible functions that estimate performance for dynamic predictions: BS, integrated BS (IBS) and \textit{tv}-AUC. In Section 2.1, we describe these functions in more detail. After estimation of $\omega_k(t)$, SL can estimate \eqref{eq:dynpred} by fitting the base learners on the complete dataset to estimate $\pi_j(u \mid t, \mathcal{M}_k)$, and then weighing these dynamic survival probabilities by the estimated $\hat{\omega}_k(t)$. \\
\indent A critical issue in SL is specifying $\mathcal{K}$. According to the philosophy of SL, users should add all possible model specifications to $\mathcal{K}$ because increasing the library size can only improve the accuracy of SL due to its oracle property. The oracle property states that as $n$ and $\mathcal{K}$ approach infinity, the eSL performs asymptotically as well as the oracle model (OM), which is the model that would perform best in an independent test dataset. The OM is typically unknown in observational studies such as the Global PBC Study, showing the property's desirability. \\
\indent Finally, in the setting of dynamic predictions, two more interlinked hyperparameters affect the performance of SL: the prediction window $\{t, u\}$ and the number of folds $V$. Stable estimation of the predictive accuracy metrics requires a sufficient number of events within the prediction window in all folds to differentiate between event-havers and non-event-havers. If the amount of events in a fold is too low, it becomes difficult for SL to determine which base learner performs best due to the unstable estimates. If the number of events is lower, dividing the data in a fewer folds, where each fold has approximately the same number of events likely leads to more stable results compared to a larger number of folds. To help achieve stable results, we create folds stratifying on the event outcome.\supercite{strat_cv} \\
\indent After specifying $\lambda(\cdot)$, $\mathcal{K}$, $\{t, u\}$ and $V$, there are two types of SL to choose from. First is the eSL where we optimally weigh the $K$ models by minimising $\lambda(\cdot)$ as shown in \eqref{eq:esl}. Second is the discrete SL (dSL), in which we choose $\mathcal{M}_k$ with the lowest cross-validated average $\lambda(\cdot)$. In other words, the dSL is the base learner that performed best in the validation set. An added benefit of working with SL for dynamic predictions is that the weights become time-dependent and can be optimised for any prediction window $\{t, u\}$. For example, in a hypothetical scenario where model A gives the most accurate predictions from $0 \le t < 10$, and model B gives the most accurate predictions for $t \ge 10$, SL will properly reflect these changes when optimising predictive accuracy in the different prediction windows. 
\subsection{Loss functions for Dynamic Predictions}
To estimate $\omega_k(t)$ for the eSL, we use the BS, IBS and \textit{tv}-AUC. Other possible options as loss functions are the expected predictive cross-entropy and mean absolute error. We focus on the BS, IBS and \textit{tv}-AUC because they are regularly used to evaluate the accuracy of dynamic predictions. A challenge in the estimation of these three predictive accuracy measures is censoring. In particular, we focus on subjects at risk at time $t$ and calculate the accuracy measures at $u > t$. We will account for censored observations in the interval $(t, u)$ using inverse probability of censoring weighting. These weights are estimated by
\begin{equation}\label{eq:ipcw}
W_i(u, t) = \frac{\mathbbm{1}(t < T_i \le u)\delta_i}{G(T_i \mid t)} + \frac{\mathbbm{1}(T_i > u)}{G(u \mid t)},
\end{equation}
where $G(\cdot)$ is the distribution for censoring times. We estimate $G(\cdot)$ using the Kaplan-Meier estimator, assuming independent censoring. In \eqref{eq:ipcw}, we see that patients who experience an event within the prediction window are weighed by the inverse probability of being censored at $T_i$. Patients who experience an event after the prediction window are weighed by the inverse probability of being censored at $u$, and patients censored in the interval are given zero weight. \\
\indent The BS measures predictive accuracy for time-to-event outcomes by balancing model discrimination and calibration.\supercite{bs} For dynamic predictions, the BS is estimated by
\begin{equation*}
\text{BS}(u \mid t) = \frac{1}{n_{\mathcal{R}(t)}} \sum_{i \in \mathcal{R}(t)} \hat{W}_i(u, t) \Bigl\{ \mathbbm{1}(T_i \le u) - \tilde{\pi}^{v}_i(u \mid t) \Bigl\} ^2,
\end{equation*}
where $n_{\mathcal{R}(t)}$ is the number of patients in $\mathcal{R}(t)$ and $\tilde{\pi}^{v}_i(u \mid t)$ is the estimate of \eqref{eq:esl}. The BS estimates error only at the end of the prediction window. If prediction error over the whole prediction interval has more medical relevance, we can estimate the IBS approximated by Simpson's rule:\supercite{simpin}
\begin{equation*}
\begin{aligned}
\text{IBS}(u \mid t) &= \frac{1}{u - t} \int_t^u \text{BS}(s, t) \, ds \\
&\approx \frac{2}{3} \text{BS}( \frac{t + u}{2} \mid t) + \frac{1}{6} \text{BS}(u \mid t).
\end{aligned}
\end{equation*} \\
\indent Lastly, the \textit{tv}-AUC measures only model discrimination. The metric is interpreted as the proportion of individuals in the prediction window correctly ranked in terms of survival probability had no one been censored, i.e., the proportion of event-havers who have a lower predicted survival probability than non-event-havers. The \textit{tv}-AUC for prediction window $\{t, u\}$ is estimated through the formula
\begin{equation*}
\text{\textit{tv}-AUC}(u,t) = \frac{\sum_{i \in \mathcal{R}(t)} \sum_{j \in \mathcal{R}(t)} \mathbbm{1}\bigl(\hat{\tilde{\pi}}_i^{v}(u,t) > \hat{\tilde{\pi}}_j^v(u,t)\bigl) D_i(u,t) \bigl(1 - D_j(u,t)\bigl) \hat{W}_i(u,t) \hat{W}_j(u,t)}{\sum_{i \in \mathcal{R}(t)} \sum_{j \in \mathcal{R}(t)} D_i(u,t) \bigl(1 - D_j(u,t)\bigl) \hat{W}_i(u,t) \hat{W}_j(u,t)},
\end{equation*} 
where $D_i(u, t) = \mathbbm{1} (t < T_i \le u) \delta_i$. In contrast with the BS and IBS, we maximise the \textit{tv}-AUC for estimation of $\omega_k(t)$. 
\subsection{Base learners for dynamic predictions}
We include base learners from three procedures in the model library of SL. The first approach we use is landmarking, a computationally simple method to obtain dynamic predictions. The concept behind landmarking is to subset $\mathcal{R}(t)$ from $\mathcal{D}_n$. Then, a time-to-event model is fit to $\mathcal{R}(t)$, where time zero denotes landmark time $t$. In landmarking, a function $f(\cdot)$ of the longitudinal outcome $\boldsymbol{y}_i$ is added to the time-to-event model as a baseline covariate. For example, if we landmark with a Cox's Proportional Hazards model\supercite{cox} (notated as Cox model or Cox regression from here on forth), the hazard function is:
\begin{equation*}
h_i(t) = h_0(t) \exp \bigl\{ \boldsymbol{\gamma}^\top \boldsymbol{\mathit{w}}_i + \sum_{m = 1}^{M} \alpha_m \mathit{f} (\boldsymbol{y}_{im} ) \bigl\},
\end{equation*}
where the baseline hazard $h_0(\cdot)$ is estimated non-parametrically, and $\boldsymbol{\gamma}$ denotes the coefficients of the baseline covariates $\boldsymbol{\mathit{w}}_i$, and $\alpha_m$ is the coefficient of function $f(\cdot)$ of longitudinal outcome $\boldsymbol{y}_{im}$. Multiple versions of landmarking exist, with a different amount of complexity and accuracy. The most basic version is one-stage landmarking, where $f(\cdot)$ denotes the last value available of $\boldsymbol{y}_{im}$ which is carried forward (LVCF) , i.e.:
\begin{equation*}
f(\boldsymbol{y}_{im}) = \mathit{y}_{m,ip}, \hspace{0.3cm} p = \max \bigl\{l \cdot \mathbbm{1}(t_{m,il} \le t) \bigl\}.
\end{equation*}
One-stage landmarking has multiple issues, however. For example, with LVCF, we assume there is no measurement error. Furthermore, we assume that the risk of an event depends only on the last observed value, and we ignore the time difference between the last measurement and the landmark time and its impact on the longitudinal outcome. One-stage landmarking has been extended to two-stage landmarking to compensate for these issues. In two-stage landmarking, we take a two-step approach. First, a linear mixed model is fit to model the longitudinal outcome. The general form of a linear mixed model for continuous $\boldsymbol{\mathit{y}}_{im}$ is
\begin{equation}\label{eq:longmod}
E (\mathit{y}_{im} (t) \mid \boldsymbol{\mathit{b}}_{im}) = \eta_{im}(t) = \boldsymbol{\mathit{x}}^\top_{im}(t) \boldsymbol{\beta}_{m} + \boldsymbol{\mathit{z}}^\top_{im}(t) \boldsymbol{\mathit{b}}_{im}, \quad \boldsymbol{\mathit{b}}_{im} \sim \mathcal{N}(\boldsymbol{0}, \boldsymbol{D}_m),
\end{equation}
where $\eta_{im}(t)$ is the expected latent trajectory of the $m$th longitudinal outcome for individidual $i$ over time, and $\boldsymbol{\mathit{x}}_{im}(t)$ and $\boldsymbol{\mathit{z}}_{im}(t)$ denote time-varying design vectors for the fixed effects $\boldsymbol{\beta_m}$ and random effects $\boldsymbol{\mathit{b}}_{im}$, respectively. In the second step, the linear mixed model is used to get an estimate of $\eta_{im}$ at $t$ to include in the two-staged landmark model:
\begin{equation*}
f(\boldsymbol{y}_{im}) = \hat{\eta}_{im}(t).
\end{equation*}
For all Cox regression landmark models, we estimate \eqref{eq:dynpred} using the Breslow estimator\supercite{breslow}
\begin{equation*}
\hat{\pi}_{j}(u \mid t) = \exp \left\{ - \hat{H}_0(u) \exp \biggr[ \boldsymbol{\gamma}^\top \boldsymbol{\mathit{w}}_j + \sum_{m = 1}^{M} \alpha_m \mathit{f} (\boldsymbol{y}_{jm} ) \biggr] \right\},
\end{equation*}
in which
\begin{equation*}
\hat{H}_0(u) = \sum_{i \in \mathcal{R}(t)} \frac{\mathbbm{1}(T_i \le u)\delta_i}{\sum_{j \in \mathcal{R}(t)} \exp\{ \boldsymbol{\gamma}^\top \boldsymbol{\mathit{w}}_j + \sum_{m = 1}^{M} \alpha_m \mathit{f} (\boldsymbol{y}_{jm}) \}}.
\end{equation*}
Note that more extensions have been made to landmark models to further improve their predictive accuracy.\supercite{lm2,lm3,lm4} Furthermore, any time-to-event landmark model can be added as a base learner to the SL ensemble (e.g., a parametric accelerated failure time landmark model). However, we adhere to Cox regression one-stage and two-stage landmarking in the current article. \\
\indent The second approach for base leaners we employ is a generalisation of a standard Cox landmark regression: a tree-based gradient-boosted Cox regression with landmarking. Gradient boosting is another form of ensemble learning, and by including these models as base learners in our SL ensemble we perform nested ensemble learning. For clarity, we denote the base learners used within gradient-boosted models as sub-base learners. A tree-based gradient-boosted Cox regression overcomes the issue of requiring a-priori information on covariate inclusion (baseline and longitudinal), model terms and interactions. The standard Cox landmark model also assumes a linear relationship between the logarithm of the hazard and the covariates, whereas a gradient-boosted version can estimate this relationship non-parametrically. \\
\indent Compared to SL, gradient boosting arrives at a final model in a different manner. In SL, the ensemble is achieved through a one-time optimisation problem, shown in \eqref{eq:esl}. In tree-based gradient boosting of a Cox model with landmarking, we arrive at an optimal model $\mathcal{M}_k \in \mathcal{K}$ by minimising the partial log-likelihood solely for $i \in \mathcal{R}(t)$ through an iterative process with the landmark time as time zero. First, we initialise our tree-based prediction function relying on $\boldsymbol{X} = (\boldsymbol{X}_1, \dots, \boldsymbol{X}_{n_{\mathcal{R}(t)}})$ with $\hat{\psi}_0(\boldsymbol{X})$. Here, $\boldsymbol{X}_i$ is short-hand notation to denote the baseline and longitudinal variables for individual $i$ at the landmark time and $\boldsymbol{X}_{n_{\mathcal{R}(t)}}$ denotes that we only look at individuals included in $\mathcal{R}(t)$. Note that the longitudinal outcomes in $\boldsymbol{X}$ can be derived through either one-stage or two-stage landmarking. Subsequently, in each iteration $b = 1, \dots, b_{stop}$, the negative gradient vector $\boldsymbol{u}_b = (u_{1,b}, \dots u_{n_{\mathcal{R}(t)},b})$ is computed. For individual $i$ in iteration $b$, the value of the negative gradient vector is derived by
\begin{equation*}
u_{i, b} = \delta_{i} - \sum_{l \in \mathcal{R}(t)} \delta_{l} \frac{\exp\Bigr\{ \hat{\psi}_{b - 1}(\boldsymbol{X}_{l}) \Bigr\}}{\sum_{k \in \mathcal{R}(t)} \exp\Bigr\{ \hat{\psi}_{b - 1}(\boldsymbol{X}_{k})\Bigr\}},
\end{equation*}
where $\hat{\psi}_{b-1}(\cdot)$ is the estimated tree-based prediction function of the boosting algorithm for iteration $b - 1$. 

Afterwards, the sub-base learners predict $\boldsymbol{u}_b$. The predicted $\hat{\boldsymbol{u}}_b$ from the sub-base learner that best predicts $\boldsymbol{u}_b$ according to the least squares estimator is then used to update the prediction function: $\hat{\psi}_b(\boldsymbol{X}) = \hat{\psi}_{b - 1}(\boldsymbol{X}) + \nu \hat{\boldsymbol{u}}_{b}$, resulting in data-driven covariate selection. The update is partly decided by $\nu$, a step-size parameter determining the proportion of the sub-base learner added to the ensemble in each iteration. For a more detailed description of the gradient boosting algorithm, see Hofner et al.\supercite{GB} \\ 
\indent There are three important hyperparameters when designing a gradient boosting model: the sub-base learners, $\nu$ and $b_{stop}$. We include only regression trees of baseline and longitudinal variables and two-way interactions as sub-base learners to model the event outcome non-parametrically. We chose a small value $(0.025)$ for $\nu$ to decrease the chances of surpassing the global minimum of the negative partial log-likelihood. Finally, we used $5$-fold cross-validation nested in $\mathcal{D}_n^{-v}$ to determine the optimal number of iterations, $b_{opt}$, to decrease the chances that the model either stops too early or overfits. \\
\indent After we fit a landmark tree-based gradient boosted Cox model, we estimate $\pi_j(u \mid t)$ with the Breslow estimator:
\begin{equation*}
\hat{\pi}_{j}(u \mid t) = \exp \biggr\{ - \hat{H}_0(u) \exp \Bigr[ \hat{\psi}_{b_{opt}}(\boldsymbol{X}_j) \Bigr] \biggr\},
\end{equation*}
in which
\begin{equation*}
\hat{H}_0(u) = \sum_{i \in \mathcal{R}(t)} \frac{\mathbbm{1}(T_i \le u)\delta_i}{\sum_{j \in \mathcal{R}(t)} \exp \Bigr\{ \hat{\psi}_{b_{opt}}(\boldsymbol{X}_j) \Bigr\}}.
\end{equation*}
Compared to a standard Cox landmark regression, the equation within the exponent is a now black-box function of the included covariates and two-way interactions. Boosted landmark models were fit using the \textbf{mboost} package.\supercite{mboost} \\
\indent The final framework we use to estimate \eqref{eq:dynpred} is joint modelling. In this framework, sub-models are defined for the two outcomes and linked through shared parameters to specify the joint distribution. For the longitudinal outcomes $\boldsymbol{\mathit{y}}_i$, a general form for the sub-model is defined in \eqref{eq:longmod}. Although the form is identical, the estimation of the longitudinal model differs in joint modelling compared to landmarking because a joint model estimates the bivariate association of the longitudinal and time-to-event outcomes. Specifically, for the time-to-event process, the joint modelling framework assumes that the instantaneous risk of having an event depends on baseline covariates, functions of the $m \in M$ latent trajectories $\eta_{im}(t)$ of the longitudinal outcomes and the vectors of random effects $\boldsymbol{\mathit{b}}_{i} = (\boldsymbol{\mathit{b}}_{i1}, \dots, \boldsymbol{\mathit{b}}_{iM})$:
\begin{equation}\label{eq:survmod}
\begin{aligned}
 h_i\{t \mid \mathcal{H}_i(t), \boldsymbol{\mathit{w}}_i\} &= \lim_{\Delta t \to 0} \frac{\text{Pr} \bigr\{ t \le T_i^* < t + \Delta t \mid T_i^* \ge t, \mathcal{H}_{i1}(t), \dots, \mathcal{H}_{iM}(t), \boldsymbol{\mathit{w}}_i \bigr\}}{\Delta t} \\
 &= h_0(t) \exp \biggr[ \boldsymbol{\gamma}^\top \boldsymbol{\mathit{w}}_i + \sum_{m = 1}^{M} \sum_{q=1}^{Q} \mathit{f}_{mq} \Bigr\{ \eta_{im}(t), \boldsymbol{\mathit{b}}_{im}, \alpha_{mq} \Bigr\} \biggr] ,
 \end{aligned}
\end{equation}
where $\mathcal{H}_{im}(t) = \{ \eta_{im}(s), 0 \le s < t \}$ represents the history of the longitudinal process up to $t$ for the $m$th longitudinal outcome, $h_0(\cdot)$ is the baseline hazard function, and $\boldsymbol{\mathit{w}}_i$ are baseline covariates with regression coefficients $\boldsymbol{\gamma}$. Function $\mathit{f}_{mq}(\cdot)$ and its parameterisation $\alpha_{mq}$ is the functional form of the $m$th longitudinal outcome and identifies its relation to the time-to-event outcome. Common examples of the functional form are the longitudinal outcome's current value, lagged value, slope or integral or a combination of multiple. For a more detailed overview of functional forms, see Rizopoulos et al.\supercite{FF} \\
\indent As evident in \eqref{eq:longmod} and \eqref{eq:survmod}, the two sub-models are linked through the distributions of random effects $\boldsymbol{\mathit{b}}_{i}$. The drawbacks of this method are that joint models assume conditional independence given $\boldsymbol{\mathit{b}}_{i}$, and estimation of the joint distribution requires integration over the random effects. Joint models are, therefore, a computationally intensive method, especially as the dimensionality of the longitudinal outcomes and random effects increases. \\
\indent Under the Bayesian joint modelling framework, $\pi_j(u \mid t)$ can be estimated with Markov chain Monte Carlo methods from the posterior predictive distribution:
\begin{equation*}
\pi_j(u \mid t) = \int \text{Pr}( T^*_j \ge u \mid T^*_j > t, \boldsymbol{\mathit{y}}_{j}(t), \boldsymbol{\theta}) \, p(\boldsymbol{\theta} \mid \mathcal{D}_n) \, d \boldsymbol{\theta},
\end{equation*}
where $\boldsymbol{\theta}$ represents all model parameters. For a more detailed explanation of dynamic prediction estimation under the joint modelling framework, see Rizopoulos et al.\supercite{dynpred} Joint models were fit using the \textbf{JMbayes2} package.\supercite{jmbayes2}
\section{Global PBC Data}
We return to our motivating example of PBC, where we apply SL to the Global PBC Data to obtain dynamic survival prediction ensembles. The Global PBC Data results from an international effort by 15 liver centres as a part of the Global PBC Study Group. We work with a subset of the complete database that contains follow-up data on 1884 PBC patients from 1988 to 2013. All patients received UDCA therapy with varying doses at the start of follow-up. Our time-to-event outcome of interest is a composite endpoint of liver decompensation, liver transplantation and death. During follow-up, 326 patients $(17.3 \%)$ experienced the composite endpoint. \\
\indent For patients, ten liver-related biomarkers were sporadically collected during follow-up time. We focus on three of the biomarkers that were measured most often, namely total bilirubin (TB), alkaline phosphatase (ALP), and aspartate aminotransferase (AST), as previous research has shown these to be strongly associated with event outcomes in PBC patients.\supercite{biomark1, biomark2} Altogether, there were 23,102 longitudinal measurement times. Patients had a rounded average of $10$ measurements for TB $(SD = 11.10)$, $11$ measurements for ALP $(SD = 11.17)$ and $11$ measurements for AST $(SD = 11.10)$. Our goal is to utilise the information in these three biomarkers and baseline covariates sex $(90.1\% \, \text{Female})$, age at inclusion $(M = 52.78, \, SD = 11.91)$, disease duration before inclusion $(M = 0.70, \, SD = 1.60)$, whether lesions were discovered in a liver biopt at inclusion $(85.0\% \, \text{Yes})$ and the year of study inclusion $(M = 2000, \, IQR = 1995 \text{-} 2005)$ to create a highly accurate and flexible dynamic prediction model for the composite endpoint in PBC patients through SL. \\ 
\indent In our library, we consider nine base learners. Models $\mathcal{M}_1$ and $\mathcal{M}_2$ are one-stage landmark models based respectively on a standard and boosted Cox regression. Model $\mathcal{M}_3$ is a standard Cox two-stage landmark model, where the linear mixed models have linear specifications for time with random intercepts and slopes. Model $\mathcal{M}_4$ is the boosted counterpart of $\mathcal{M}_3$. Because the three biomarkers have been shown to follow nonlinear trajectories,\supercite{biomark1} we included another standard Cox two-stage landmark model with nonlinear time specification in the longitudinal model through natural cubic splines as model $\mathcal{M}_5$. We also included a boosted version of $\mathcal{M}_5$ with model $\mathcal{M}_6$. Finally, we included three joint models based on clinical knowledge from hepatologists: a joint model with linear subject-specific time trends assuming current value association between all biomarkers and hazard ($\mathcal{M}_7$), a joint model with nonlinear subject-specific time trend and current value association between all biomarkers and hazard ($\mathcal{M}_8$), and a joint model with nonlinear subject-specific time trend, where the assumed association between the biomarkers and hazard were: current value and slope for TB, current value and area for ALP, and current value and area for AST ($\mathcal{M}_9$). For an overview of the included models in the model library, we refer readers to the Appendix. \\
\indent PBC is a slow progressive chronic disease. Therefore, we estimated cross-validated predictions in two prediction windows with high clinical relevance. The prediction windows we chose are year 6 to 9 and year 9 to 12 since study inclusion. At year 6, $1257$ patients were at risk of experiencing the event. During years 6 to 9, $77$ patients experienced the composite event, while $373$ patients were censored. At year 9, $807$ patients remained in the risk set. In the following three years, another $43$ patients experienced the event while 287 patients were censored. \\
\indent We used the following procedure to train the SL. First, we created an independent test set that contained 20\% of the total data. Then, we estimate the SL for the two prediction windows on the remaining data (i.e., training data) using a cross-validation procedure. Specifically, we split the training data into seven and four folds for the prediction windows $\{6, 9\}$ and $\{9, 12\}$ respectively to have enough events in the time intervals in all folds of the validation set. Then, we used the three performance metrics described in Section 2.1 to estimate model weights for the eSL and determine the dSL. After estimation of the eSL and dSL, we train the base learners on the complete training data. In total, the two prediction windows contain 61 and 31 events. However, landmark-based base learners are trained on all events past the chosen landmark time. This means that all landmark models were trained on respectively 137 and 76 events for the two prediction windows, allowing us to learn more complex relations from the data. Finally, we estimated the performance metrics of the fitted models separately, the eSL and the dSL in the independent test set and compared the results. For a more complete overview of the data analysis, we included a pseudo-algorithm in the Appendix. \\
\indent We present the results of the analysis in Table 1 and Figure 1. Table 1 shows the time-dependent property of SL weights for dynamic predictions, indicating that changing the weights of models lead to a better performance in different prediction windows. For example, we find that for year $9$ to $12$, the weight for $\mathcal{M}_8$ under the IBS increases compared to year $6$ to $9$, indicating that the $\mathcal{M}_8$ most likely performed better for the later prediction window. \\ 
\begin{table}[t!]
	\parbox{0.5\linewidth}{
	\centering
		\begin{tabular}{ l | c c c }
			\toprule
			\multicolumn{1}{c}{} & \multicolumn{3}{c}{Loss function} \\
			 Model & BS & IBS & \textit{tv}-AUC \\
			 \toprule
			1 & \textbf{0.00} & \zero & \zero \\
			2 & \zero & \zero & 0.07 \\
			3 & \zero & \zero & 0.02 \\
			4 & 0.47 & 0.23 & \textbf{0.15} \\
			5 & \zero & \zero & \zero \\
			6 & \zero & \zero & \zero \\
			7 & 0.35 & \textbf{0.77} & 0.27 \\
			8 & \zero & \zero & 0.46 \\
			9 & 0.18 & \zero & 0.03 \\
			\midrule
			\bottomrule	
	\end{tabular}
}
	\hfill
	\parbox{0.5\linewidth}{
	\centering
 	\begin{tabular}{ l | c c c }
                    \toprule
                    \multicolumn{1}{c}{} & \multicolumn{3}{c}{Loss function} \\                     
                    Model & BS & IBS & \textit{tv}-AUC \\
                    \toprule
                    1 & \textbf{0.89} & \zero & 0.13 \\
                    2 & \zero & \zero & \textbf{0.00} \\
                    3 & \zero & \zero & \zero \\
                    4 & \zero & \zero & 0.30 \\
                    5 & \zero & \zero & 0.32 \\
                    6 & \zero & \zero & 0.05 \\
                    7 & 0.11 & \zero & 0.06 \\
                    8 & \zero & \textbf{1.00} & 0.14 \\
                    9 & \zero & \zero & 0.00 \\
                    \midrule
                    \bottomrule
	\end{tabular}
}
\caption{\small{Super Learner weight estimates at prediction window \{6, 9\} (left) and \{9, 12\} (right) under the three different loss functions described in Section 2.1 for predicting the composite endpoint in the Global PBC Data. The boldfaced numbers denote the discrete SL for a given prediction window and loss function. Models $\mathcal{M}_1$ to $\mathcal{M}_9$ are denoted by 1 to 9.}}
\end{table}
\indent Finally, we noted that when optimising the \textit{tv}-AUC, we often ran into convergence issues. We argue this occurred due to the nature of the \textit{tv}-AUC as a proportion, where the gradient of the loss function quickly becomes flat. This leads to optimisation algorithms getting stuck at local maxima when maximising the \textit{tv}-AUC. We combatted this issue by running the optimisation algorithm multiple times, each time using different starting values. Afterwards, we chose the optimised weights that resulted in the highest \textit{tv}-AUC. \\
\indent Inspecting the predictive performance of SL and the base leaners in the independent test set, Figure 1 shows that the eSL tends to perform as well as or better than any model fit separately on the different predictive accuracy metrics, displaying the benefit of SL. This is because the best model depends on both the metric used to assess performance and the specified time window. However, with SL we always optimise predictive accuracy, allowing high predictive accuracy without putting too much emphasis into one model being correctly specified. The results display only point estimates of the performance because confidence intervals for SL for dynamic predictions can only be achieved through bootstrapping the whole SL procedure, which is practically challenging to implement due to the large size of the data and the computational complexity of the more complex base learners. For an impression of the variance of the performance of SL for dynamic predictions, we refer readers to the simulation results in Section 4.2.
\begin{figure}[t!]
	\begin{center}
	\includegraphics[width=0.9\textwidth]{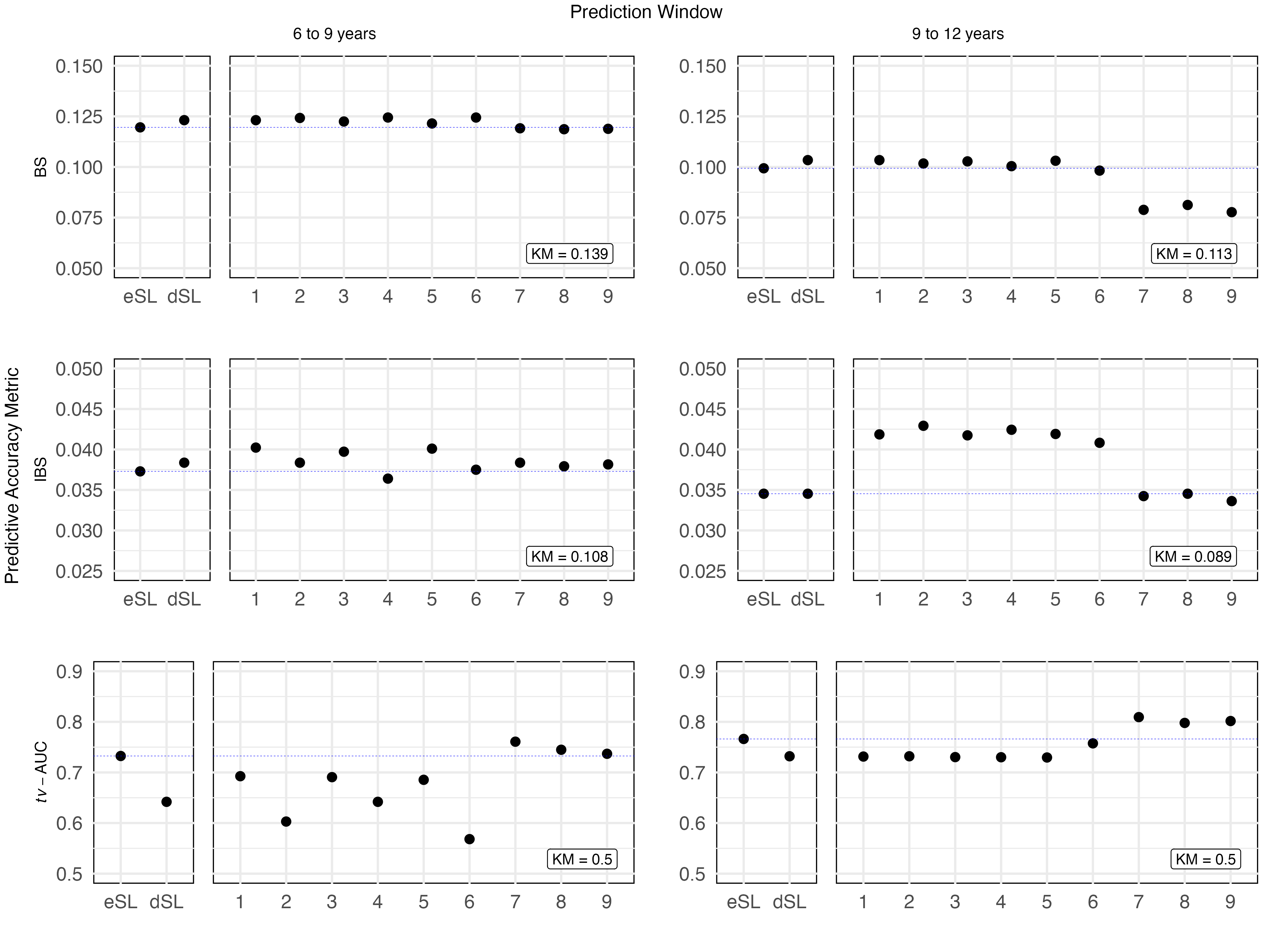}
	\end{center}
\caption{\small{Metric estimates of the ensemble SL (eSL) and discrete SL (dSL) for predicting the composite endpoint in the Global PBC Data. Models $\mathcal{M}_1$ to $\mathcal{M}_9$ are denoted by $1$ to $9$. The blue dashed line serves as reference point for the performance of the eSL. The textboxes in each graph denote the performance of a Kaplan-Meier (KM) model that contained no covariates. }}
\end{figure}
\section{Simulation Study}
In Section 2, we mention that we estimated the different predictive accuracy metrics using Kaplan-Meier to estimate the IPCW. However, using Kaplan-Meier for IPCW ignores a frequent issue in the setting of both longitudinal and time-to-event outcomes: the issue of informative censoring. Although there is no strong indication of informative censoring in the Global PBC Data, where most patients were censored due to administrative censoring, it often occurs that a treating physician may remove patients from a study for their safety if their condition deteriorates. In these cases, the decision to censor a subject from the study depends on their biomarker values. Misspecification of the censoring distribution by using Kaplan-Meier will then bias error estimation. Kvamme and Borgan\supercite{inf_cens} discuss this issue in more depth and have shown that ignoring informative administrative censoring may result in over-optimistic BS. As a result, we may give more weight to a model's predictions, thinking it performs better than it actually does. Therefore, we perform a simulation study to examine the impact of informative censoring on the performance of SL and its estimates of the three predictive accuracy metrics.
\subsection{Simulation Design}
We designed a simulation study to examine the impact of informative censoring on the performance of SL in the setting of longitudinal and time-to-event data. We tried to emulate a setting where there is no single true model in the following way. For the longitudinal process, we generated data from one process: a mixed model with a linear time trend and parameters based on the Global PBC dataset with measurement times taken from a uniform distribution. For the survival process, we generate data from two different models. The first model imitates a one-stage landmark process. At each measurement time, we generate event times based on the latest recorded longitudinal value and baseline variables using the method from Bender et al.\supercite{bender}. If the event time is shorter than the next measurement time, we assume the individual experienced the event and terminate the collection of additional longitudinal measurements. If the event time exceeds the next measurement time, we simulate a new event time based on the subsequent longitudinal outcome. In the second setting, we generate data from a joint model process using inverse transform sampling. Specifically, for each person, we estimate the hazard function in continuous time based on the longitudinal process and baseline covariates. Then by integrating out over time, we simulate event times for each individual by deriving at which point in time the cumulative hazard equals their draw from a standard uniform distribution.  \\
\indent After generating longitudinal and event data from the two processes $(n = 625)$, we censor under three different scenarios. In Scenario 1, we use the data as is - i.e., no censoring. In Scenario 2, we generate censoring times from a uniform distribution and censor those whose censoring time is lower than their event time to model random censoring. In Scenario 3, we assume an informative censoring process. At each measurement time point, we estimate the probability of being censored from a logistic distribution based on the current longitudinal value and baseline covariates. To ensure a fair comparison, we controlled the number of individuals censored to be similar within the prediction window between Scenario 2 and 3. Finally, we randomly sampled $125$ individuals from the total dataset to form the independent test set, leaving 500 observations for fitting SL. We ran the simulation $100$ times. \\
\indent For each scenario, we used 5-fold cross-validation to train SL with three base learners: the true one-stage landmark model and the true joint model, described above, and a two-stage landmark model. For all base learners, we employed the same longitudinal model. After estimating the weights for the eSL, we compared the performance of the eSL with the performance of the dSL and the OM - i.e., the model that performed best in the independent test set. In all scenarios, we estimated IPCW using Kaplan-Meier for the censoring distribution.
\subsection{Results}
Figure 2 presents the results of the estimates of the various metrics in the 100 simulated training and test datasets of the three scenarios. Table 2 shows the average SL weights and standard deviation for the different loss functions under the three scenarios. \\
\begin{figure}[t!]
	\begin{center}
	\includegraphics[width=0.90\textwidth]{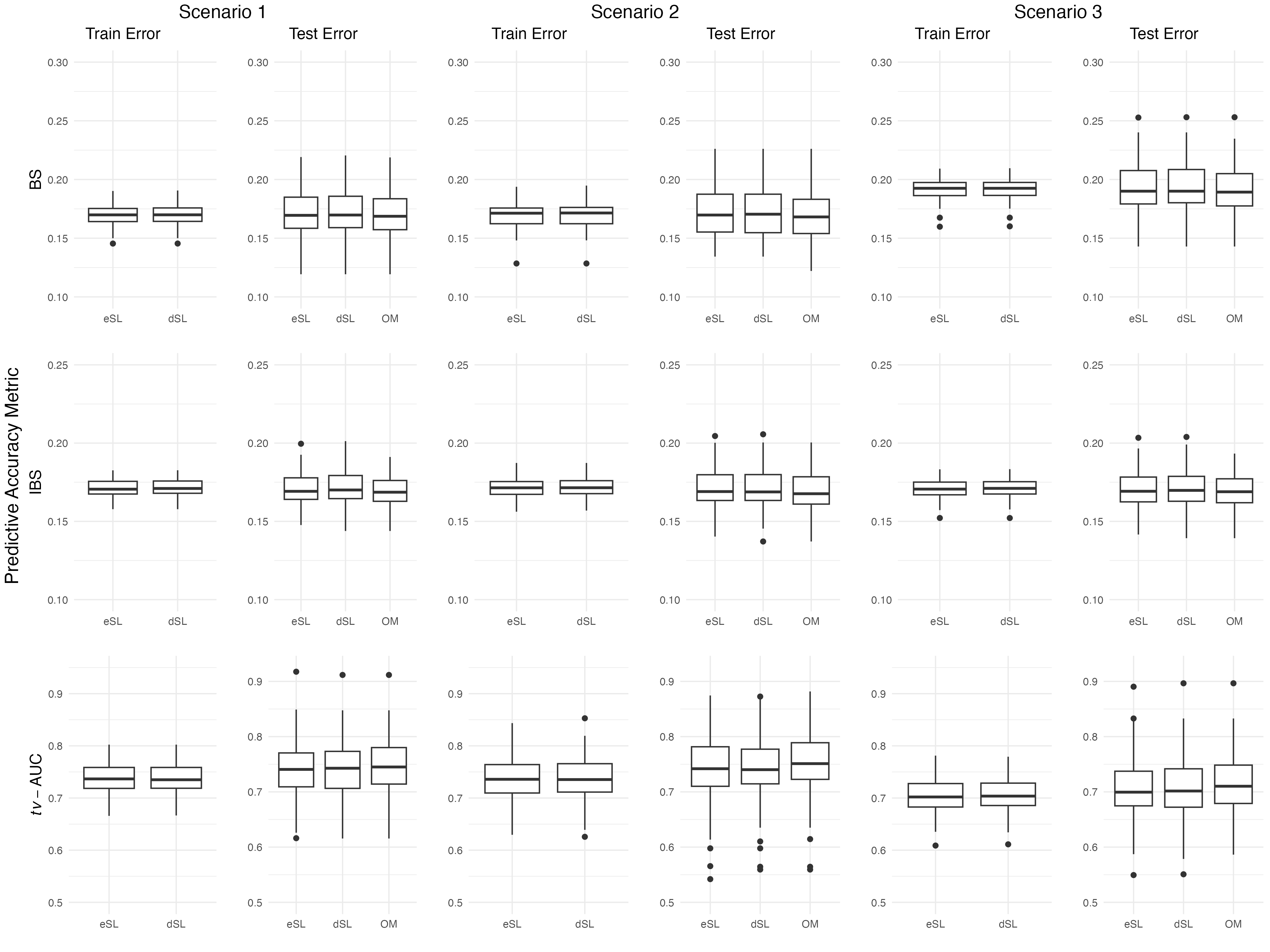}
	\end{center}
	\caption{\small{Metric estimates of the ensemble SL (eSL) and discrete SL (dSL) and oracle model (OM) under no censoring (Scenario 1), random censoring (Scenario 2) and informative censoring (Scenario 3) over 100 simulated datasets. The box plots under Train Error denote the metric values in the dataset the model was trained on, the box plots under Test Error show the metric values in the holdout data set.}}
\end{figure}
\indent For the \textit{tv}-AUC and BS, we find the expected result that the estimated performance under Scenario 2 is close to the true performance in Scenario 1 in both the training and test set. Comparing Scenario 1 and Scenario 3, we find that the BS increases and \textit{tv}-AUC decreases in Scenario 3. That is to say, inappropriately accounting for informative censoring leads to a decrease in predictive accuracy, according to the BS and \textit{tv}-AUC. These results differ from those shown of Kvamme and Borgan.\supercite{inf_cens} A possible explanation for the different findings is a dissimilarity in censoring mechanisms. For example, had we imposed stronger informative censoring in Scenario 3, more individuals would be censored before the prediction window. Consequently, there would be fewer events in the prediction window, ensuring that the BS in Scenario 3 is lower than in Scenario 2. \\
\indent Looking at the distribution of the weights for the BS and \textit{tv}-AUC, the two true models receive the most weight. However, joint models tend to receive the larger weight. Joint models might dominate the weights for two reasons. First, they are more efficient than the landmarking method, as joint models can use all the longitudinal data up to the prediction window versus just the last observed value. Second, there is a loss of information in the landmark model as it cannot model the time gap between the last measurement time and the start of the prediction window. \\ 
\indent We find different results for the IBS. First, compared to the BS and \textit{tv}-AUC the weights for the IBS seem to shift more toward the true landmark model than the joint model, indicating that the landmark model has lower estimated error at the halfway point of the prediction window. However, a more important difference is that the IBS under Scenario 2 does not seem to differ from the error in Scenario 3. We hypothesise that this happens because we estimate the censoring distribution at two time points by approximating the IBS through Simpson's rule. The differences between the two censoring mechanisms might only begin to show more prominently at later time points, which is weighed down in Simpson's rule. From this perspective, the IBS gives biased results towards the impact of censoring on the model's performance, ignoring informative censoring. \\
\setlength{\tabcolsep}{2pt}
\begin{table}[t!]
	\centering
 	\begin{tabular}{l | c c c | c c c | c c c}
                    \toprule
                    \multicolumn{1}{c}{} & \multicolumn{3}{c}{Scenario 1} & \multicolumn{3}{c}{Scenario 2} & \multicolumn{3}{c}{Scenario 3} \\                     
                    Model & BS & IBS & \textit{tv}-AUC & BS & IBS & \textit{tv}-AUC & BS & IBS & \textit{tv}-AUC \\
                    \toprule
                    1 & 0.30 (0.32) & 0.47 (0.28) & 0.34 (0.20) & 0.23 (0.29) & 0.35 (0.31) & 0.37 (0.23) & 0.31 (0.33) & 0.56 (0.30) & 0.36 (0.24) \\
                    2 & 0.00 (0.00) & 0.04 (0.05) & 0.01 (0.04) & 0.01 (0.02) & 0.04 (0.06) & 0.01 (0.05) & 0.00 (0.01) & 0.02 (0.04) & 0.01 (0.05) \\
                    3 & 0.79 (0.32) & 0.50 (0.28) & 0.64 (0.20) & 0.76 (0.29) & 0.61 (0.32) & 0.62 (0.23) & 0.69 (0.33) & 0.42 (0.29) & 0.63 (0.23) \\
                   
                    \midrule
                    \bottomrule
	\end{tabular}
\caption{\small{Average model weight and its standard deviation in brackets under no censoring (Scenario 1), random censoring (Scenario 2) and informative censoring (Scenario 3). Model 1 is the true one-stage landmark model, model 2 is a two-stage landmark model and model 3 is the true joint model.}}
\end{table}
\indent Investigating the performance of SL in the different scenarios, we see that the performance of the dSL and eSL remains close to that of the OM for the different predictive accuracy metrics. These results indicate that the oracle property of SL also holds under non-random censoring mechanisms for dynamic predictions. \\
\indent Finally, we checked convergence for the eSL due to the issues we encountered for the \textit{tv}-AUC in the Global PBC Data. Convergence was a greater issue for the \textit{tv}-AUC, where SL did not properly converge $12.0\%$ of the time over all scenarios. In comparison, the proportion of non-convergence for the BS and IBS were $0.00$ and $0.02$, respectively. These proportion further strengthen the importance of caution when optimising the \textit{tv}-AUC.
\section{Discussion}
In the current article, we further extended the Super Learner framework in the longitudinal and time-to-event data setting. Our extensions combine the predictive performance of different dynamic prediction frameworks that work in continuous time. We also included a new predictive accuracy metric for the framework, the \textit{tv}-AUC. Applying our method to optimise predictive accuracy for patients suffering from PBC using the Global PBC data, we found that the eSL performed as well as or better than any single model on the different predictive accuracy metrics with more flexibility in model specification. In our simulation study, we showed that under informative censoring, the performance of the eSL came close to the performance of the OM in an independent test set. However, informative censoring negatively impacts the performance of models, which the IBS may not correctly take into account. \\
\indent The SL framework presented offers unique and valuable advantages to the setting of longitudinal and time-to-event outcomes due to the large variability in procedures and model specification. 
For future use cases of SL for dynamic predictions, we recommend researchers to create a model library that is tailored to their objective. If computational efficiency is a primary concern, then the library should consist mostly of landmark models and simple boosting algorithms as these are fastest to fit. However, if the objective is optimising predictive accuracy, it is worth creating a model library that is as broad and diverse as possible while keeping computational efficiency in mind. In the creation of the model library, it is also worth considering the base learners separately and as a whole. That is to say, it may be worth adding a relatively simple base learner that by itself does not predict well yet predicts something else than the other base learners. Furthermore, previous research in similar research settings can be used to help guide the creation of the model library. For example, variable importance measures from previous studies can help guide covariate inclusion for the base learners. Nonetheless, it is important to note that the flexibility and accuracy of SL are not without costs, as SL decreases model interpretation. We caution against over-interpreting model weights and model coefficients from individual base learners.
Future extensions of the framework include allowing even more diverse methods in the model library, such as discretised-time methods, and appropriately changing the loss functions to the context of competing risks. We also found that informative censoring negatively impacted the predictive accuracy of SL. Therefore, another avenue of future research is to investigate how we can mitigate the impact of informative censoring on the predictive performance of SL. \\
\indent Finally, the framework we presented is fully predictive. A future direction we are interested in is counterfactual prediction to answer causal questions such as: what would the dynamic survival probability of a patient with an inadequate response on UDCA therapy have been, had they received an add-on therapy of obeticholic acid? Van der Laan and Polley\supercite{SL} developed SL as a semiparametric method to obtain an initial estimate of the probability distribution. This initial estimate can then be updated with extra assumptions to get a targeted causal estimate through Targeted Learning.\supercite{TL} Extending Targeted Learning to the setting of longitudinal and time-to-event data would be a highly relevant addition to drawing possible causal conclusions from observational longitudinal and time-to-event data.
%
%
%
\printbibliography

\newpage
\section*{Appendix}
\section*{Global PBC Data}
We present two tables and two pseudo-algorithms that detail the analysis of the Global PBC Data using Super Learner for dynamic predictions of time-to-event outcomes. Table 3 shows the packages and their version number that we used for the different types of base learners that we included in the ensemble and their hyperparameters. Table 4 presents a summary of all included base learners in the ensemble. Finally, Algorithms 1 and 2 detail the fitting and estimation process of the base learners and Super Learner for dynamic predictions. \\
\newpage
\setcounter{table}{2}
\begin{table}[h]
\centering
	\begin{tabular}{ l | c r c }
    	\toprule
	    \multicolumn{1}{l}{} & \multicolumn{3}{c}{Details} \\
	    Base learner & Package & Version & Hyperparameters \\
	    \toprule
	    Landmark: Standard Cox & survival & 3.6-4 & -\\
	    Landmark: Boosted Cox & mboost & 2.9-10 & $b_{stop}$, $\nu$, $b_{opt}$ \\
	    Joint Model & JMbayes2 & 0.5-0 & priors, MCMC samples, chains, burn in \\
	    \midrule
	    \bottomrule
	\end{tabular}
\caption{\small{Details of the implementation of the different type of base learners and their hyperparameters.}}
\end{table}
\indent For all boosted models, $b_{stop}$ was set to $500$, $\nu$ was set to 0.025 and the early stopping criteria $b_{opt}$ was determined using $5$-fold cross-validation in the training set (see Algorithm 2). The tree-based learners in the boosting algorithm also contain many hyperparameters, such as the minimum split size and the maximum depth of the trees. For all tree-related hyperparameters, we used the $\textbf{mboost}$ package defaults. For joint models, we used the $\textbf{JMbayes2}$ package defaults for all priors. For the fitting parameters, we employed two MCMC chains with a total number of 5000 iterations of which  1000 were burn-in. For more complete information on the fitting process for the SL, we refer readers to Algorithm 1 and the Code and Data Supplement. \\
\newpage
\begin{table}[h]
\centering
	\begin{tabular}{ c | c c c | c c c}
    	\toprule
	    \multicolumn{1}{l}{} & \multicolumn{3}{c}{Details} & \multicolumn{3}{c}{Functional forms} \\
	    Base learner & Type & Algorithm & Longitudinal model & TB & ALP & AST \\
	    \toprule
	    $\mathcal{M}_1$ & One-stage LM & Standard Cox & - & -& - & -\\
	    $\mathcal{M}_2$ & One-stage LM & Boosted Cox & - & -& - & -\\
	    $\mathcal{M}_3$ & Two-stage LM & Standard Cox  & linear & -& - & -\\
	    $\mathcal{M}_4$ & Two-stage LM & Boosted Cox  & linear & -& - & -\\
	    $\mathcal{M}_5$ & Two-stage LM & Standard Cox  & splines (df = 3) & -& - &-\\
	    $\mathcal{M}_6$ & Two-stage LM & Boosted Cox  & splines (df = 3 & - & - & - \\
	    $\mathcal{M}_7$ & JM &-  & linear & CV & CV & CV\\
	    $\mathcal{M}_8$ & JM & - & splines (df = 3) & CV & CV & CV \\
	    $\mathcal{M}_9$ & JM & -  & splines (df = 3) & CV \& S & CV \& A & CV \& A \\
	    \midrule
	    \bottomrule
	\end{tabular}
\caption{\small{Summary table for the $9$ base learners included in the Super Learner ensemble. The column Type details the type of base learner, Algorithm defines how the base learner is estimated, Longitudinal model details the time specification for the base learner, and Functional form shows the association between per longitudinal outcome and the hazard function. A dash implies non-applicability. The abbreviations are LM = landmark; JM = joint model; df = degrees of freedom; CV = current value; S = slope; A = area.}}
\end{table}
\newpage

\begin{algorithm}
\caption{Pseudo-Algorithm of Super Leaner for Dynamic Predictions Application}
\begin{algorithmic}[1]
\State \textbf{Initialise}
\State Specify $\lambda(\cdot)$, $\mathcal{K}$, $\{t, u\}$, $V$\
\State Split $\mathcal{D}_n$ into $\mathcal{D}_n^{\text{train}}$ and $\mathcal{D}_n^{\text{test}}$

\State
\State \textbf{Base learners}
\State Split $\mathcal{D}_n^{\text{train}}$ up in $V$ folds, stratifying on event indicator $\delta_i$
\For{$v$ in $1:V$}
    \State Train $\mathcal{M}_k \in \mathcal{K}$ on $\mathcal{D}_n^{\text{train},-v}$:
    \For{$\mathcal{M}_k \in \mathcal{K}$}
    	\If{$\mathcal{M}_k$ = landmark model}
        		\State Run Algorithm 2 (see below)
    	\ElsIf{$\mathcal{M}_k$ = joint model}
		\State {Define mixed model functions $\eta_{im}(t)$ for $m = 1, \dots, M$}
		\State {Define hazard function $h_i\{t \mid \mathcal{H}_i(t), \boldsymbol{\mathit{w}}_i\} = h_0(t) \exp \biggr[ \boldsymbol{\gamma}^\top \boldsymbol{\mathit{w}}_i + \sum_{m = 1}^{M} \sum_{q=1}^{Q} \mathit{f}_{mq} \Bigr\{ \eta_{im}(t), \boldsymbol{\mathit{b}}_{im}, \alpha_{mq} \Bigr\} \biggr]$ }
		\State {Estimate $\eta_{im}(t)$ and $h_i\{t \mid \mathcal{H}_i(t), \boldsymbol{\mathit{w}}_i\}$ simultaneously}
	\EndIf
	\State Predict $\pi_i^v(u \mid t, \mathcal{M}_k)$
    \EndFor
\EndFor
\State Return $\hat{\pi}_i^v(u \mid t, \mathcal{M}_k)$ for all $i$ in $\mathcal{R}(t)$ and all $\mathcal{M}_k \in \mathcal{K}$

\State
\State \textbf{Meta learner}
\State Estimate $\hat{\omega}_k(t) = \underset{\omega_k(t)}{\text{argmin}} \biggr\{ \sum_{i \in \mathcal{R}(t)} \lambda \Bigr( \sum_{k = 1}^K \omega_k(t) \hat{\pi}_i^v(u \mid t, \mathcal{M}_k),\, T_i \Bigr) \biggr\},$ \\ subject to $\omega_k(t) \ge 0 \ \wedge \ \sum_{k = 1}^{K} \omega_k(t) = 1$

\State
\State \textbf{Evaluate}
\State Train all $\mathcal{M}_k \in \mathcal{K}$ on full $\mathcal{D}_n^{\text{train}}$
\State Evaluate $\tilde{\pi}_j^v(u | t) = \sum_{k = 1}^K \hat{\omega}_k(t) \hat{\pi}_j^v(u \mid t, \mathcal{M}_k)$ on $\mathcal{R}(t)^{\text{test}} = \{ j \in \mathcal{D}_n^{\text{test}} \mid T_j > t \}$

\end{algorithmic}
\end{algorithm}

\newpage

\begin{algorithm}
\caption{Fitting Process for Landmark Base Learners}
\begin{algorithmic}[1]
\State Filter $\mathcal{R}(t)^{-v} = \{ i \in \mathcal{D}_n^{\text{train},-v} \mid T_i > t\}$
\State
\If{$\mathcal{M}_k$ = one-stage landmark model}
	\State{For all $i \in \mathcal{R}(t)^{-v}$, $f(\boldsymbol{y}_{im}) = \mathit{y}_{m,ip}$, where $p = \max \bigl\{l \cdot \mathbbm{1}(t_{m,il} \le t) \bigl\}$ for $m = 1, \dots, M$} 
	\ElsIf{$\mathcal{M}_k$ = two-stage landmark model}
		\State Filter $\boldsymbol{y}_{i}(t)$ for all $i \in \mathcal{D}_n^{\text{train},-v}$
		\State Define and estimate mixed model functions $\eta_{im}(t)$ for $m = 1, \dots, M$
		\State For all $i \in \mathcal{R}(t)^{-v}$, $f(\boldsymbol{y}_{im}) = \hat{\eta}_{im}(t)$ for $m = 1, \dots, M$
\EndIf
\State
	\If{$\mathcal{M}_k$ = standard Cox model}
		\State Fit Cox's Proportional Hazards Model on $\mathcal{R}(t)^{-v}$, where $h_i(t) = h_0(t) \exp \bigl\{ \boldsymbol{\gamma}^\top \boldsymbol{\mathit{w}}_i + \sum_{m = 1}^{M} \alpha_m f(\boldsymbol{y}_{im}) \bigl\}$
	\ElsIf{$\mathcal{M}_k$ = boosted Cox model}
		\State \textbf{Boosting}
		\State Create tree-based learners for all covariates and two-way interactions in $\boldsymbol{X}$ for $i \in \mathcal{R}(t)^{-v}$
		\State Initialise prediction function $\psi_0(\boldsymbol{X})$
		
		\For{$b$ in $1:b_{stop}$}
			\State Compute negative gradient vector $\boldsymbol{u}_b$
			\State Predict $\boldsymbol{u}_b$ with tree-based learners
			\State Select best predicted $\hat{\boldsymbol{u}}_b$ according to least squares criteria
			\State Update the estimated prediction function: $\hat{\psi}_b(\boldsymbol{X}) = \hat{\psi}_{b - 1}(\boldsymbol{X}) + \nu \hat{\boldsymbol{u}}_b$
		\EndFor
		\State
		\State \textbf{Early stopping}
		\State Split $\mathcal{R}(t)^{-v}$ into $p = 1, \dots, 5$ folds
		\For{$p$ in $1:5$}
			\State Filter individuals from fold $p$ from $\mathcal{R}(t)^{-v}$ to create $\mathcal{R}(t)^{-v,-p}$
			\State Repeat \textbf{Boosting}, where $\mathcal{R}(t)^{-v} = \mathcal{R}(t)^{-v,-p}$
			\State Estimate the negative partial log likelihood of the model on the individuals in fold $p$		
		\EndFor
		\State{Determine $b_{opt}$ = $b$ where the average negative partial log likelihood over folds is lowest}
		\State{Return $\hat{\psi}_{b_{opt}}(\boldsymbol{X})$ on $\mathcal{R}(t)^{-v}$}
	\EndIf
\end{algorithmic}
\end{algorithm}

\end{document}